\documentclass{llncs}

\usepackage{amsmath,amssymb,amsfonts,amscd,array,bm}
\usepackage{multirow}
\usepackage{graphicx}
\usepackage{listings}
\usepackage{nicefrac}
\usepackage{enumerate}

\usepackage{subfigure} 

\usepackage{hyperref}

\def\muacro#1{\texttt{#1}}          				

\def\eqlabel#1{\label{eq:#1}}
\def\eqref#1{Equation~(\ref{eq:#1})}
\def\eqsref#1#2{Equations~(\ref{eq:#1},~\ref{eq:#2})}

\def\seclabel#1{\label{sec:#1}}
\def\secref#1{Section~\ref{sec:#1}}
\def\secsref#1#2{Sections~\ref{sec:#1},~\ref{sec:#2}}

\def\tablabel#1{\label{tab:#1}}
\def\tabref#1{Table~\ref{tab:#1}}

\def\thlabel#1{\label{th:#1}}
\def\thref#1{Theorem~\ref{th:#1}}
\def\thsref#1#2{Theorems~\ref{th:#1},~\ref{th:#2}}



\def\vect#1{{\bm{#1}}}

\def\normal#1{\mathcal{N}\left(#1\right)}

\def\distGeneral#1#2#3{\delta_{#1}\left(#2,#3\right)}
\def\VSP#1{\vect{\mathcal{#1}}}

\def\argmax{\operatornamewithlimits{arg\,max}}
\def\argmin{\operatornamewithlimits{arg\,min}}

\begin{document}
\frontmatter          
\pagestyle{headings}  
\addtocmark{Intrinsic Dimension} 

\title{\muacro{DANCo}: Dimensionality from Angle and Norm Concentration}
\titlerunning{\muacro{DANCo}: Dimensionality from Angle and Norm Concentration}

\author{C. Ceruti \and S. Bassis \and A. Rozza \and
G. Lombardi \and E. Casiraghi \and P. Campadelli}
\authorrunning{C.Ceruti et al.}
\tocauthor{C. Ceruti, S. Bassis, A. Rozza, G. Lombardi, E. Casiraghi, and P. Campadelli}
\institute{Dipartimento di Scienze dell'Informazione, Universit\`a degli Studi di Milano, \\ 
					 Via Comelico 39-41, 20135 Milano, Italy \\
					 \email{claudio.ceruti@unimi.it}}
\maketitle              

\begin{abstract}
In the last decades the estimation of the intrinsic dimensionality of a dataset has gained considerable importance. 
Despite the great deal of research work devoted to this task, 
most of the proposed solutions prove to be unreliable when the intrinsic dimensionality of the input dataset is high and the manifold where the points lie is nonlinearly embedded in a higher dimensional space. In this paper we propose a novel robust intrinsic dimensionality estimator that exploits the twofold complementary information conveyed both by the normalized nearest neighbor distances and by the angles computed on couples of neighboring points, providing also closed-forms for the Kullback-Leibler divergences of the respective distributions.
Experiments performed on both synthetic and real datasets highlight the robustness and the effectiveness of the proposed algorithm when compared to state of the art methodologies.
\keywords{Intrinsic dimensionality estimation, manifold learning, von Mises distribution, Kullback-Leibler divergence.}
\end{abstract}

\section{Introduction}\seclabel{Introduction}

Given a dataset $\vect{X}_N\equiv\{\vect{x}_i\}_{i=1}^N\subset\Re^D$,  
its intrinsic dimension (\muacro{id}) is the minimum number of parameters 
needed to represent the data without information loss. 
In the last decade a great deal of research work has been devoted to the 
development of \muacro{id} estimation algorithms; to this aim,
the feature vectors $\vect{x}_i$ are generally 
viewed as points constrained to lie on a low dimensional manifold $\VSP{M} \subseteq \Re^d$ 
embedded in a higher dimensional space $\Re^D$, where $d$ is the \muacro{id} to be estimated.
In more general terms, according to~\cite{Fukunaga1982}, $\vect{X}_N$ is said 
to have \muacro{id} equal to $d \in \{1..D\}$ if its elements lie entirely within a $d$-dimensional subspace of $\Re^D$.

The \muacro{id} is a very useful information for the following reasons. 
At first, dimensionality reduction techniques, which are often used to reduce the ``curse of dimensionality'' effect~\cite{Bellman1961} by computing a more compact representation of the data, are profitable when the number of projection dimensions is the minimal one that allows to retain the maximum amount of useful information expressed by the data.
Furthermore, when using an auto-associative neural network~\cite{Kirby2001} to perform a nonlinear feature extraction, the \muacro{id} can suggest a reasonable value for the number of hidden neurons.
Moreover, according to the statistical learning theory~\cite{Vapnik1998}, the capacity and the generalization capability of
a classifier may depend on the \muacro{id}.
In particular, in~\cite{Friedman2009} the authors mark that, in order to balance a classifier's generalization ability and its empirical error, the complexity of the classification model should also be related to the \muacro{id} of the available dataset.
Finally, as it has been recently shown in~\cite{Camastra2009}, \muacro{id} estimation methods are used to evaluate the model order in a time series, which is crucial to make reliable time series predictions; this consideration is supported by the fact that the domain of attraction of a nonlinear dynamic system has a very complex geometric structure and the studies on the geometry of the attraction domain are closely related to fractal geometry, and therefore to fractal dimension.

Unfortunately, even if a great deal of research work has been focused at the development of \muacro{id} estimators, and several interesting techniques have been presented in the literature, to our knowledge only few methods~\cite{Camastra2002,ECML2011,ICIAP2011,journalMlj} have investigated the problem of input datasets having a sufficiently high \muacro{id} (that is \muacro{id} $\geqslant 10$) and being drawn from manifolds nonlinearly embedded in higher dimensional spaces; 
this fact is also highlighted by the experiments reported in this paper showing that well-known techniques fail when dealing with this kind of data. 
More precisely, it can be noted that several methods underestimate the \muacro{id} if its value is too high.
These considerations lead us to the development of an \muacro{id} estimator, called ``\muacro{DANCo}'' (Dimensionality from Angle and Norm Concentration), that is less affected by underestimation problems, as it is shown by experiments on both synthetic and real datasets, and by the comparison of the achieved results with those reported by state of the art algorithms.
The peculiarities and strengths of the proposed estimator are to be sought in the joint use of normalized nearest neighbor distances and mutual angles, whose coupled exploitation allows to reduce the effects of well-known problems such as curse of dimensionality, edge effect, and overall orthogonality. 

\smallskip
This paper is organized as follows: 
in \secref{RelWork} previous works on \muacro{id} estimators are reviewed. In \secref{ThResults} base theoretical results laying foundations for the proposed estimator are presented. \secref{algorithm} sketches the proposed algorithm providing a concise analysis of its properties. A detailed comparison with state of the art methodologies on a wide family of datasets is summarized in \secref{Experimental}. Finally, \secref{Conclusions} reports conclusions and future works. 

\section{Related Works}\seclabel{RelWork}

In this section we summarize the literature related to \muacro{id} estimation methods; note that a more detailed description is reported in the survey~\cite{Camastra2003}.
\par \smallskip

The most cited \muacro{id} estimator is the Principal Component Analysis (\muacro{PCA})~\cite{Jollife1986}, which projects the input dataset on the $d$ directions of maximum variance (principal components, \muacro{PC}s). 
Exploiting \muacro{PCA}, the estimated \muacro{id} is the number of \muacro{PC}s whose corresponding normalized eigenvalues are higher than a thresholding parameter, usually difficult to be set.
More accurate results can be obtained by applying a local \muacro{PCA}~\cite{Fukunaga1971} that determines
the \muacro{id} by combining local estimates computed in small subregions of the dataset; 
unfortunately, complications arise in the identification of local regions and in the selection of thresholds~\cite{Verveer1995}.
In~\cite{Bishop1998} Bishop describes a Bayesian treatment of \muacro{PCA} (\muacro{BPCA}) to automatically estimate the \muacro{id} of the input dataset. 
This technique has been extended in~\cite{Li2010} to cope with exponential family distributions, but this method requires the knowledge of the distribution underlying the data.
To achieve an automatic selection of meaningful \muacro{PC}s, in~\cite{Guan2009} the authors propose the Sparse Probabilistic Principal Component Analysis (\muacro{SPPCA}) that exploits the sparsity of the projection matrix through a probabilistic Bayesian formulation. \muacro{PCA}-based methods, such as those previously mentioned, are usually classified as projection methods~\cite{Camastra2003,Levina2005} since they search for the best subspace where to project the data; unfortunately, they cannot provide reliable \muacro{id} estimates since they are too sensitive to noise and parameter settings~\cite{Levina2005}. 

\smallskip
Geometric \muacro{id} estimation methods~\cite{Levina2005} are most often based on some statistics related to either the distances between neighboring points or the fractal dimension, expressing them as functions of the \muacro{id} of the embedded manifold. 
The most popular fractal dimension estimator is the Correlation Dimension~(\muacro{CD})~\cite{Grassberger1983} that is based on the assumption that the volume of a $d$-dimensional set scales as $r^d$ with its size $r$.
Since the performances of \muacro{CD} are affected by the choice of the scale $r$, in~\cite{Hein2005} the author suggests an 
estimator (here called~\muacro{Hein}) based on the asymptotes of a smoothed version of the \muacro{CD} estimate. 
In~\cite{Farahmand2007} the authors present an algorithm to estimate the 
\muacro{id} of a manifold in a small neighborhood of a selected point, and they analyze its finite-sample convergence properties.
Another technique, based on the analysis of point neighborhoods, is the Maximum Likelihood Estimator~(\muacro{MLE})~\cite{Levina2005} that applies the principle of maximum likelihood to the distances between neighboring points. 
In~\cite{Costa2004} the authors propose an algorithm that exploits entropic graphs to estimate both the 
\muacro{id} and the intrinsic entropy of a manifold;
they test their method by adopting either the Geodesic Minimal Spanning Tree (\muacro{GMST}~\cite{Costa2004b}), 
where the arc weights are the geodetic distances computed through the \muacro{ISOMAP} algorithm~\cite{Tenenbaum2000}, 
or the more efficient \muacro{kNN}-graph~(\muacro{kNNG}~\cite{Costa2004}), where the arc weights are based on Euclidean distances. 

\smallskip
We note that many neighborhood based estimators usually underestimate the \muacro{id} when its value is sufficiently high and, to our knowledge, only few works address this problem~\cite{Camastra2002,ICIAP2011,ECML2011}.
Indeed, as shown in~\cite{Eckmann1992}, the number of sample points required to perform dimensionality estimation grows exponentially with the \muacro{id} (``curse of dimensionality''). For this reason, 
when the dimensionality is too high the number of sample points practically available is insufficient to compute an acceptable \muacro{id} estimation. 
Moreover, the ratio between the points close to the edge of the manifold and the points inside it raises in probability when the dimensionality increases (``edge effect'',~\cite{Verveer1995}), affecting the results achieved by estimators based on statistics related to the behavior of point neighborhoods.

In~\cite{Camastra2002}, the authors propose an empirical \muacro{id} correction procedure based on the estimation 
of the error obtained on synthetically produced datasets of known dimensionality. More precisely, after generating 
$D$ datasets characterized by incremental \muacro{id} values ($d_i\in\{1..D$\}), the authors apply the \muacro{CD} algorithm~\cite{Grassberger1983} to estimate the \muacro{id} ($\hat{d}_i$) of each dataset. Fitting the points $(d_i,\hat{d}_i)$ they obtain the so-called ``correction curve'' used to adjust the \muacro{id} estimates.
In~\cite{ICIAP2011} a local estimator (called \muacro{IDEA}) based on an asymptotic correction is proposed.  
To this aim, given a dataset of unknown \muacro{id}, random subsets of different cardinalities are extracted and their \muacro{id} estimates are computed; the bi-dimensional points composed by the cardinality of each subset and by its \muacro{id} estimate are then fitted with a curve having a horizontal asymptote whose ordinate is the final \muacro{id}. 
In~\cite{ECML2011} the authors describe a method (called $\muacro{MiND}_\muacro{KL}$) based on the comparison between the empirical probability density function of the neighborhood distances computed on the dataset and the distribution of the neighborhood distances computed from points uniformly drawn from hyperspheres of known increasing dimensionality; the \muacro{id} estimate is the one minimizing the Kullback-Leibler divergence (\muacro{KL}).

\section{Theoretical Results}\seclabel{ThResults}

Consider a manifold $\VSP{M}\equiv\Re^d$ embedded in a higher dimensional space $\Re^D$ through a locally isometric nonlinear smooth map $\phi:\Re^d\to\Re^D$; to estimate the \muacro{id} of $\VSP{M}$ by means of points drawn from the embedded manifold through a smooth probability density function (\muacro{pdf}) $f$, we need to identify a ``mathematical object'' depending only on $d$, and we should define a consistent estimator for $d$ based on it.

Assume by hypothesis that the employed manifold sampling process is driven by a smooth \muacro{pdf} $f$; moreover, consider a spherical neighborhood of the origin $\vect{0}_d$ having radius $\epsilon$; 
denoting with $\chi_{\VSP{B}_d(\vect{0}_d,1)}$ the indicator function on the unit ball $\VSP{B}_d(\vect{0}_d,1)$, the \muacro{pdf} restricted to such a neighborhood is:
{\small
	\begin{equation}\eqlabel{fDelta}
		f_{\epsilon}(\vect{z}) = 
			\frac{f(\epsilon\vect{z}) \chi_{\VSP{B}_d(\vect{0}_d,1)}(\vect{z})}
				{\int_{\vect{t}\in\VSP{B}_d(\vect{0}_d,1)}f(\epsilon\vect{t})d\vect{t}}
	\end{equation}
}

In~\cite{ECML2011} the authors prove the following:
\begin{theorem}\thlabel{th1}
Given $\{\epsilon_i\}\to0^+$,~\eqref{fDelta} describes a sequence of \muacro{pdf}s having the unit $d$-dimensional ball as support; such sequence converges uniformly to the uniform distribution $\mathbf{B}_d$ in the ball $\VSP{B}_d(\vect{0}_d,1)$.
\end{theorem}

\thref{th1} ensures that, from a theoretical standpoint, in our setting it is possible to assume uniformly distributed points in every neighborhood of $\VSP{M}$; 
in other words, it is possible to define consistent estimators based on local information, assuming without loss of generality that the normalized points are uniformly drawn from the unit hypersphere. 

Our technique exploits the statistical properties of norms and mutual angles computed on points drawn from uniformly sampled hyperspheres;
to this aim, in~\secsref{dist}{Angle} we sketch the statistical properties of norms and angles respectively, while in~\secref{together} we describe how both the above properties can be simultaneously used to define a consistent estimator of the manifold's \muacro{id}.

\subsection{Concentration of Norms}\seclabel{dist}

Consider initially the problem of estimating the \muacro{id} of $\VSP{M}$ by means of a sample $\{\vect{z}_i\}_{i=1}^k$ of points uniformly drawn from $\VSP{B}_d(\vect{0}_d,1)$; to this aim, we exploit the concentration of norms that is dimensionality-dependent.

In~\cite{ECML2011} it is shown that the \muacro{pdf} associated to the normalized distance $r$ between the hypersphere center and its nearest neighbor is the following: 
{
	\begin{equation}\eqlabel{pdfFirstNeigh}
		g(r;k,d) = kdr^{d-1}(1 - r^d)^{k-1}
	\end{equation}
}

\vspace{-0.5cm}
\thref{th1} proves that the convergence of $f_\epsilon$ to $\mathbf{B}_d$ is uniform, so that when $\epsilon\to0^+$ the \muacro{pdf} related to the geodetic distances 
$\frac{1}{\epsilon}\distGeneral{\phi}{\phi(\vect{0}_d)}{\phi(\vect{z})}$ 
converges to the \muacro{pdf} $g$ defined in \eqref{pdfFirstNeigh}.
Notice that, once $k$ is fixed, $\mathcal G=\{g(r;k,d)\}_{d=1}^D$ represents a finite family of $D$ \muacro{pdf}s for all the parameter values $1\leq d\leq D$.

\smallskip
As reported in~\cite{ECML2011}, a Maximum Likelihood estimator (\muacro{ML}) could be found for the parameter $d$ of $g$, but 
the resulting estimate may be poor due to the usage of the \muacro{kNN} algorithm. 
More precisely, in high dimensional spaces, the \muacro{kNN} method is strongly affected by the edge effect~\cite{Verveer1995} that reduces the quality of the neighborhood estimation.

To obtain a more reliable estimate of $d$, in~\cite{ECML2011} the authors propose to minimize the \muacro{KL} divergence between the \muacro{pdf} computed on the dataset and those calculated on synthetic data of known \muacro{id}s; to this aim, they adopted the \muacro{KL} method proposed in~\cite{Wang2006}. 

However, though this \muacro{KL} approach can be applied to every dataset without any restriction on the underlying distribution,
in our problem the closed-form for the \muacro{KL} divergence between two minimum neighbor distance \muacro{pdf}s can be analytically identified.
To this aim, once the parameter $k$ is fixed, we need to estimate the parameter $d$ in $g$; to accomplish this task, we decided to employ the \muacro{ML} estimator proposed in~\cite{ECML2011}. Calling $\hat{d}_{ML}$ the \muacro{ML} estimation obtained on the dataset, and $\check{d}_{d,ML}$ the \muacro{ML} estimations obtained by means of points sampled from $d$-dimensional hyperspheres\footnote{Notice that, due to the \muacro{kNN} bias effect described above, the \muacro{ML} estimates $\check{d}_{d,ML}$ are biased w.r.t.\ the real value $d$ employed in the sampling process, and a similar bias can be observed also in the estimated $\hat{d}_{ML}$.} (for $d\in\{1..D\}$), the closed-form of the \muacro{KL} for the minimum neighbor distances is:
{\footnotesize
	\begin{multline}\eqlabel{KLnorms}
		\overline{KL}_{d}
			= \VSP{KL}(g(\cdot;k,\hat{d}_{ML}),g(\cdot;k,\check{d}_{d,ML})) 
			= \int_0^1 g(r;k,\hat{d}_{ML})\log\left(\frac{g(r;k,\hat{d}_{ML})}{g(r;k,\check{d}_{d,ML})}\right) \mathrm{d}r \\
			= \VSP{H}_k\frac{\check{d}_{d,ML}}{\hat{d}_{ML}} - 1 - \VSP{H}_{k-1} - \log{\frac{\check{d}_{d,ML}}{\hat{d}_{ML}}}
				-(k-1)\sum_{i=0}^k (-)^i \binom{k}{i}\Psi\left(1+\frac{i\hat{d}_{ML}}{\check{d}_{d,ML}}\right) 
	\end{multline}}%
where $\VSP{KL}(\cdot,\cdot)$ is the \muacro{KL} divergence operator, $\VSP{H}_k$ represents the $k$-th harmonic number $\left(\VSP{H}_k=\sum_{i=1}^k\frac{1}{i}\right)$, and $\Psi(\cdot)$ is the digamma function.

\subsection{Concentration of Angles}\seclabel{Angle}

As it happens for norms, in high dimensions pairwise angles among $k$ uniformly distributed unitary vectors $\{\vect{x}_i\}_{i=1}^k$ on a $(d-1)$-dimensional surface $S^{d-1}$ of a hypersphere in $\Re^d$ are subject to the concentration of their values.
The common belief that in high dimensions such vectors tend to be orthogonal to each other has found partly theoretical justification in the past~\cite{Mardia1972}, but only in the last decades an even deeper investigation has allowed a more precise characterization of this fact~\cite{Sodergren2011}.

Two of the most common distributions adopted in circular and directional statistics are the von Mises distribution (\muacro{VM}) and its high-dimensional generalization termed von Mises-Fisher distribution (\muacro{VMF}). 
More precisely, for $\vect{x} \in S^{d-1}$, the \muacro{VMF} distribution has the following form: 
\begin{equation}\eqlabel{eqPdfVMF}
	q(\vect{x}; \vect{\nu}, \tau, d) = 
			C_{d}(\tau)\exp \left( {\tau \vect{\nu}^T \vect{x} } \right)
\end{equation}%
where the unit vector $\vect{\nu}$ denotes the mean direction, and the concentration parameter $\tau \ge 0$ gets high values in case of a high concentration of the distribution around the mean direction. In particular, $\tau=0$ when points are uniformly distributed on $S^{d-1}$.
Moreover, the normalization constant $C_{d}(\tau)$ in \eqref{eqPdfVMF} takes the following form:
\begin{equation}\nonumber
	C_{d}(\tau) = \frac {\tau^{d/2-1}} {(2\pi)^{d/2}I_{d/2-1}(\tau)}
\end{equation}%
where $I_{v}$ is the modified Bessel function of the first kind with order $v$.
Due to the normalization factor, this \muacro{pdf} is difficult to be used in theoretical derivations; moreover, in the assumptions of 
\thref{th1}, no information about $d$ may be estimated by the knowledge of parameters $(\vect{\nu},\tau)$, being $\vect{\nu}$ uninformative when the hyphersolid angles are uniformly distributed ($\tau=0$), which is the case of a uniformly sampled hypersphere.

Therefore, to infer the \muacro{id} of $\VSP{M}$ by exploiting the angular information, we focused on the distribution of the angles $\theta$ computed between independent pairs of random points chosen in neighborhoods of $\Re^d$ and sampled from the uniform distribution in the hypersphere. 
Note that working on pairwise angles allows both to exploit the concentration factor $\tau$, which is strictly related to the dimensionality $d$ as we will show, and to rely on the \muacro{VM} distribution, which is more tractable w.r.t. the \muacro{VMF} \muacro{pdf}.

With the above notation, considering the angle $\theta\in[-\pi,\pi]$ between two vectors, the \muacro{VM} \muacro{pdf} of $\theta$ reads as: 
\begin{equation}\eqlabel{eqPdfVM}
	q(\theta;\nu,\tau) =
		\frac{\mathrm e^{\tau\cos(\theta-\nu)}}{2\pi I_0(\tau)}\chi_{[-\pi,\pi]}(\theta)
\end{equation}%
with the same parameters and notation adopted for the \muacro{VMF} \muacro{pdf}. Intuitively, the \muacro{VM} distribution is the circular counterpart of the normal distribution on a line, sharing with the latter many interesting properties~\cite{Breitenberger1963}.
To understand the link between $\tau$ and $d$, we firstly recall that $q(\theta;\nu,\tau)$ is unimodal for $\tau>0$, as a Gaussian random variable peaked around its mean.
Next, according to the following theorem, increasing values of $\tau$ are expected for points uniformly drawn from hyperspheres with increasing dimensionality $d$.
\begin{theorem}
\thlabel{thmAngle}Given two independent random unit vectors $(\vect x_1,\vect x_2)$ in $\Re^d$, chosen from a uniform distribution on $S^{d-1}$, the concentration parameter $\tau$ of the \muacro{VM} distribution describing the angle $\theta$ between $\vect x_1$ and $\vect x_2$ converges asymptotically to the dimensionality $d$.
\begin{proof}
Consider the following results:
	\begin{enumerate}[i)]
		\item for large concentration values $\tau$, a \muacro{VM} distribution with parameters $(\nu,\tau)$ becomes a Gaussian distribution with mean $\nu$ and standard deviation $\nicefrac{1}{\sqrt{\tau}}$~\cite{Hill1976};
		\item performing the variable substitution $\tilde\theta=\sqrt{d}(\theta-\nicefrac{\pi}{2})$, the resulting random variable converges in distribution to a standard normal one~\cite{Sodergren2011}. 
	\end{enumerate}
Combining i) and ii), it follows that $\theta$ asymptotically follows a Gaussian \muacro{pdf} with mean $\nu=\nicefrac{\pi}{2}$ and standard deviation $\sigma=\nicefrac{1}{\sqrt{\tau}}=\nicefrac{1}{\sqrt{d}}$, which holds only when $\tau=d$. 
\end{proof}
\end{theorem}
\thref{thmAngle} has both a general and a specific value. At first, it formally proves the existence of the concentration of angles in high dimensions, stating both an asymptotic linear relation between concentration and dimensionality, and the orthogonality between any couple of infinite-dimensional vectors.
Moreover, \thref{thmAngle} allows to estimate the \muacro{id} ($d$) of the observed points through the estimation of the concentration parameter $\tau$.

The methodology we propose in~\secref{algorithm} employs both the \muacro{ML} estimation of the \muacro{VM} parameters $\nu$ and $\tau$, and the \muacro{KL} divergence between the \muacro{VM} \muacro{pdf} estimated from the observed dataset and those computed on synthetic data of known \muacro{id}s.
Assuming that $\{\theta_1,\ldots,\theta_N\}$ is a sample drawn from a \muacro{VM} distribution with parameters $(\nu,\tau)$, the \muacro{ML} of the population direction $\nu$ equals the sample mean direction; more precisely: 
{\small
	\begin{equation}\eqlabel{eqMLETheta}
		\hat\nu = \arctan{\frac{\sum_{i=1}^N\sin\theta_i}{\sum_{i=1}^N\cos\theta_i}}	
	\end{equation}%
}
Likewise, the \muacro{ML}  of the concentration parameter
$\tau$ equals the concentration parameter $\hat\tau$ calculated as a solution of $\eta = \frac{I_1(\tau)}{I_0(\tau)}\equiv A(\tau)$, where $\eta$ is the norm of the sample mean vector defined in~\cite{Upton1986} as:
{\small
	\begin{equation}
		\eta=\sqrt{\left(\frac{1}{N}\sum_{i=1}^N\cos\theta_i\right)^2+\left(\frac{1}{N}\sum_{i=1}^N\sin\theta_i\right)^2} 
	\end{equation}
}

Being $A$ a non invertible function, we rely on the well-known and qualified method proposed in~\cite{Fisher1981}, which approximates $A^{-1}(\eta)$ by:
{\small
	\begin{equation}\eqlabel{tau}
		\hat\tau = \widetilde A^{-1}(\eta) =
			\begin{cases}
				2\eta+\eta^3+\frac{5\eta^5}{6} & 				\quad \eta<0.53 \\
				-0.4+1.39\eta+\frac{0.43}{1-\eta} & 	\quad 0.53\leq \eta<0.85 \\
				\frac{1}{\eta^3-4\eta^2+3\eta} & 				\quad \eta\geq 0.85
			\end{cases}
	\end{equation}
}

Once an estimate of the \muacro{VM} \muacro{pdf} is obtained, we need to compare it with those computed on synthetic data of known \muacro{id}s. To this aim, 
a closed-form of the \muacro{KL} between two \muacro{VM} \muacro{pdf}s of parameters $(\nu_1,\tau_1)$ and $(\nu_2,\tau_2)$ is defined in~\cite{Oraintara2008} as:
\begin{eqnarray}\eqlabel{WMKL}
\overline{KL}_{\nu,\tau} &=& \VSP{KL}(q(\cdot;\nu_1,\tau_1),q(\cdot;\nu_2,\tau_2)) = \int_{-\pi}^{\pi} q(\theta;\nu_1,\tau_1)\log\left(\frac{q(\theta;\nu_1,\tau_1)}{q(\theta;\nu_2,\tau_2)}\right) \mathrm{d}\theta \nonumber\\
&=& \log\frac{I_0(\tau_2)}{I_0(\tau_1)} + \frac{I_1(\tau_1) - I_1(-\tau_1)}{2I_0(\tau_1)}\left(\tau_1 - \tau_2\cos(\nu_2-\nu_1)\right)
\end{eqnarray}

\subsection{Combining Angle and Norm Concentration}\seclabel{together}

In the previous sections we described the base theory laying foundations for an \muacro{id} estimator exploiting the information conveyed by the concentration of norms and angles.
To provide a unique technique that combines these information, we should compare the joint empirical \muacro{pdf} $\hat{h}(r,\theta)$ related to the real dataset with the $D$ theoretical \muacro{pdf}s, which will be referred to as $h_d(r,\theta)$ (where $d\in\{1..D\}$). Summarizing, the \muacro{id} estimate we want to compute is:
	\begin{equation}\nonumber
		\hat{d} = \argmin_{1\leq d\leq D} \int_{-\pi}^{\pi}\int_0^1 h_d(r,\theta)\log\left(\frac{h_d(r,\theta)}{\hat{h}(r,\theta)}\right) \mathrm{d}r \mathrm{d}\theta
	\end{equation}
Note that neither the theoretical $h_d$ is easily derivable, nor the joint \muacro{pdf} $\hat{h}$ can be precisely estimated. Luckily, the norm distribution $g(r;k,d)$ and the angle distribution $q(\theta;\nu,\tau)$ are independent when the data are uniformly drawn from a spherical distribution~\cite{Lord1954}; therefore the joint \muacro{pdf} factorizes in the product of the two marginals, i.e.\ $h_d(r,\theta)=g(r;k,d)q(\theta;\nu,\tau)$, so that the \muacro{KL} divergence between $h_d(r,\theta)$ and $\hat{h}(r,\theta)$ becomes:
\begin{equation}\eqlabel{KLtheoretical}
	\overline{KL}_{d,\nu,\tau} = \VSP{KL}(h_d(r,\theta),\hat{h}(r,\theta)) = \overline{KL}_{d} + \overline{KL}_{\nu,\tau}
\end{equation}

This fact allows to split the joint $\overline{KL}_{d,\nu,\tau}$ in the sum of the two closed-form divergences reported in \eqsref{KLnorms}{WMKL}; it follows that the \muacro{id} estimator exploited in our algorithm becomes: $\hat{d} = \argmin_{1\leq d\leq D} \overline{KL}_{d,\nu,\tau}$.

\section{The Algorithm}\seclabel{algorithm}

In this section we show how the theoretical results presented in~\secref{ThResults} can be exploited to estimate the \muacro{id} of a given dataset combining the information expressed by the angles and by the minimum neighbor distances.

More precisely, we consider a manifold $\VSP{M}\equiv \Re^{d}$ embedded in a higher dimensional space $\Re^{D}$ through a locally isometric nonlinear smooth map $\phi : \VSP{M} \to \Re^{D}$, and a sample set $\vect{X}_N=\{\vect{x}_i\}_{i=1}^N=\{\phi(\vect{z}_i)\}_{i=1}^N\subset\Re^D$, where $\vect{z}_i$ are independent identically distributed points drawn from $\VSP{M}$ according to a non-uniform smooth \muacro{pdf} $f:\VSP{M} \to \Re^{+}$.

To estimate the \muacro{id} of $\VSP{M}$, for each point $\vect{x}_i\in\vect{X}_N$ we find the set of $k+1$ ($1 \leq k \leq N-2$) nearest neighbors $\bar{\vect{X}}_{k+1}=\bar{\vect{X}}_{k+1}(\vect{x}_i)=\{\vect{x}_j\}_{j=1}^{k+1}\subset\vect{X}_N$.
Calling $\hat{\vect{x}}=\hat{\vect{x}}_{k+1}(\vect{x}_i)\in\bar{\vect{X}}_{k+1}$ the farthest neighbor of $\vect{x}_i$,
we calculate the distance between $\vect{x}_i$ and its nearest neighbor in $\bar{\vect{X}}_{k+1}$, and we normalize it by means of the distance between $\vect{x}_i$ and $\hat{\vect{x}}$. 
More precisely:
\begin{equation}\eqlabel{vectrho}
	\rho(\vect{x}_i) = \min_{\vect{x}_j\in\bar{\vect{X}}_{k+1}} \frac{\|\vect{x}_i - \vect{x}_j\|}{\|\vect{x}_i - \hat{\vect{x}}\|}
\end{equation}
This equation is used to compute a vector of normalized distances $\hat{\vect{r}}=\{\hat{r}_{i}\}_{i=1}^N=\{\rho(\vect{x}_i)\}_{i=1}^N$. By employing Equation~(7) in~\cite{ECML2011}, we compute the \muacro{ML} estimation by numerically solving the optimization problem $\hat{d}_{ML} = \argmax_{1\leq d\leq D}ll(d)$, where:
{\footnotesize
	\begin{equation}
		ll(d) = N\log kd + (d-1)\sum_{\vect{x}_i\in\vect{X}_N} \log \rho(\vect{x}_i) \nonumber
				        + (k-1)\sum_{\vect{x}_i\in\vect{X}_N} \log\left(1-\rho^d(\vect{x}_i)\right) \nonumber
	\end{equation}
}

Similarly, for each point $\vect{x}_i\in\vect{X}_N$ we find its $k$ nearest neighbors $\bar{\vect{X}}_{k}$ and we center them by means of a translation to obtain $\hat{\vect{X}}_{k}=\left\{ \vect{x}_{j} - \vect{x}_i :~ \forall \vect{x}_{j} \in \bar{\vect{X}}_{k} \right\}$; next, we calculate $\binom {k} {2}$ angles of all the possible pairs of vectors in $\hat{\vect{X}}_{k}$, as follows:
\begin{equation}\eqlabel{vecttheta}
	\theta(\vect{x}_z,\vect{x}_j) = \arccos \frac{\vect{x}_z \cdot \vect{x}_j}{{\|\vect{x}_z\| \|\vect{x}_j\|}}
\end{equation}
For each neighborhood we compute a vector $\hat{\vect{\theta}}=\{\theta(\vect{x}_z,\vect{x}_j)\}_{1 \leq i < j \leq k}$ by means of \eqref{vecttheta}.
Since $\hat{\vect{\theta}}$ follows a \muacro{VM} \muacro{pdf} of parameters $\nu$ and $\tau$ (see~\secref{Angle}), we estimate their values by employing the \muacro{ML} approach described in \eqsref{eqMLETheta}{tau} for each set of neighbors, thus obtaining the vectors $\hat{\vect{\nu}}=\{{\hat{\nu}_i}\}_{i=1}^N$ and $\hat{\vect{\tau}}=\{{\hat{\tau}_i}\}_{i=1}^N$, and their means $\hat{\mu}_{\nu} = N^{-1}\sum_{i=1}^N\hat{\nu}_i$ and $\hat{\mu}_{\tau} = N^{-1}\sum_{i=1}^N\hat{\tau}_i$.

Moreover, for each dimensionality $d \in \{1..D\}$ we uniformly draw a set of $N$ points $\vect{Y}_{Nd}=\{\vect{y}_i\}_{i=1}^N$ from the 
unit $d$-dimensional hypersphere\footnote{Notice that a $d$-dimensional vector randomly sampled from a $d$-dimensional hypersphere according to the uniform \muacro{pdf} can be generated by drawing a point $\bar{\vect{y}}$ from a standard normal distribution $\normal{\cdot|\vect{0}_d,1}$ and by scaling its norm.}, and we similarly compute a vector of normalized distances $\check{\vect{r}}_d=\{\check{r}_{id}\}_{i=1}^N=\{\rho(\vect{y}_i)\}_{i=1}^N$ and its \muacro{ML} estimation $\check{d}_{d,ML}$. Next, we calculate the vectors of the \muacro{VM} distribution parameters $\check{\vect{\nu}}_{d}=\{{\nu_i}\}_{i=1}^N$ and $\check{\vect{\tau}}_{d}=\{{\tau_i}\}_{i=1}^N$ together with their means $\check{\mu}_{\nu}^{d}$ and $\check{\mu}_{\tau}^{d}$.

Finally, we compose \eqsref{KLnorms}{WMKL} as reported in \eqref{KLtheoretical}, thus obtaining the following \muacro{id} estimate: 
{\small
	\begin{equation}\eqlabel{empiricalEstim}
		\hat{d} = \argmin_{d\in\{1..D\}} 
			\VSP{KL}(g(\cdot;k,\hat{d}_{ML}),g(\cdot;k,\check{d}_{d,ML})) + \VSP{KL}(q(\cdot;\hat{\mu}_{\nu},\hat{\mu}_{\tau}),q(\cdot;\check{\mu}_{\nu}^{d},\check{\mu}_{\tau}^{d}))
	\end{equation}
}
We call this \muacro{id} estimator \muacro{DANCo} (Dimensionality from Angle and Norm Concentration). Its time complexity is $O(D^2N \log N)$ and it is dominated by the time complexity of the \muacro{kNN} algorithm ($O(DN\log N)$).

Considering Theorem~4 in~\cite{Costa2005}, which ensures that geodetic distances in the infinitesimal ball converge to Euclidean distances with probability $1$, and the results in~\thsref{th1}{thmAngle}, \eqref{empiricalEstim} represents a 
consistent estimator for the \muacro{id} of the manifold $\VSP{M}$.

\section{Algorithm Evaluation}\seclabel{Experimental}
In this section we describe the datasets employed in our experiments (see~\secref{datasets}), we summarize the adopted 
experimental settings (see~\secref{expSettings}), and we report the results achieved by the proposed algorithm, comparing them to those obtained by 
state of the art \muacro{id} estimators (see~\secref{results}).

\subsection{Dataset Description}\seclabel{datasets}

To evaluate our algorithm, we have performed experiments on the $17$ synthetic and $5$ real datasets reported in~\tabref{synthDatasets}. 
In details, to generate $15$ synthetic datasets we have employed the tool proposed in~\cite{Hein2005}, extending it to 
produce the datasets $\VSP{M}_{13}$ and $\VSP{M}_{14}$ by drawing points from  
nonlinearly embedded manifolds having high \muacro{id}. More precisely, to generate $\VSP{M}_{13}$ we have proceeded as follows: 
starting from $2500$ points $\{\vect{x}_i\}_{i=1}^{2500}$ uniformly drawn in $[0,1]^{18}$, we multiplied each $\vect{x}_i$ first by $\sin(\cos(2\pi \vect{x}_i))$, then by $\cos(\sin(2\pi \vect{x}_i))$, obtaining points in $[0,1]^{36}$ after a concatenation of the above coordinates. The dataset $\VSP{M}_{13}$, containing $2500$ points in $[0,1]^{72}$, was finally obtained by duplicating each point's coordinate; this dataset, whose \muacro{id} is $18$, is composed by points drawn from a manifold nonlinearly embedded in $\Re^{72}$. 
The dataset $\VSP{M}_{14}$ was similarly generated starting from the same number of uniformly sampled points in $\Re^{24}$.

\begin{table}[!t]
	\centering
	\caption{Brief description of the $17$ synthetic and $5$ real datasets employed in our experiments, where $d$ is the \muacro{id} and $D$ is the embedding space dimension.}
	\tablabel{synthDatasets}
{
	\begin{tabular}{|c|c|c|c|l|}
		\hline
		\textbf{Dataset} & \textbf{Name} & $\vect{d}$ & $\vect{D}$ & \textbf{Description} \\
		\hline
		\hline
		\multirow{17}{*}{\textbf{Syntethic}} 
		& $\VSP{M}_{1}$ & $10$ & $11$ & Uniformly sampled sphere linearly embedded. \\ 
		& $\VSP{M}_2$ & $3$ & $5$ & Affine space. \\ 
		& $\VSP{M}_3$ & $4$ & $6$ & Concentrated figure, confusable with a $3d$ one. \\ 
		& $\VSP{M}_4$ & $4$ & $8$ & Nonlinear manifold. \\ 
		& $\VSP{M}_5$ & $2$ & $3$ & 2-d Helix \\ 
		& $\VSP{M}_6$ & $6$ & $36$ & Nonlinear manifold. \\ 
		& $\VSP{M}_7$ & $2$ & $3$ & Swiss-Roll. \\ 
		& $\VSP{M}_8$ & $20$ & $20$ & Affine space. \\ 
		& $\VSP{M}_{9a}$ & $10$ & $11$ & Uniformly sampled hypercube. \\ 
		& $\VSP{M}_{9b}$ & $17$ & $18$ & Uniformly sampled hypercube. \\ 
		& $\VSP{M}_{9c}$ & $24$ & $25$ & Uniformly sampled hypercube. \\ 
		& $\VSP{M}_{9d}$ & $70$ & $71$ & Uniformly sampled hypercube. \\ 
		& $\VSP{M}_{10}$ & $2$ & $3$ & M\"oebius band $10$-times twisted. \\ 
		& $\VSP{M}_{11}$ & $20$ & $20$ & Isotropic multivariate Gaussian. \\ 
		& $\VSP{M}_{12}$ & $1$ & $13$ & Curve. \\
		& $\VSP{M}_{13}$ & $18$ & $72$ & Nonlinear manifold. \\
		& $\VSP{M}_{14}$ & $24$ & $96$ & Nonlinear manifold. \\

		\hline
		\hline
		\multirow{5}{*}{\textbf{Real}} 
		& $\VSP{M}_\muacro{Faces}$ & $3$ & $4096$ & \muacro{ISOMAP} face dataset. \\ 
		& $\VSP{M}_\muacro{MNIST1}$ & $8-11$ & $784$ & \muacro{MNIST} database (digit $1$). \\ 
		& $\VSP{M}_\muacro{SantaFe}$ & $9$ & $50$ & \muacro{Santa Fe} dataset (version $D2$). \\ 
		& $\VSP{M}_\muacro{Isolet}$ & $16-22$ & $617$ & Spoken letter of the alphabet \\ 
		& $\VSP{M}_\muacro{DSVC1}$ & $2.26$ & $20$ & Real time series of a Chua's circuit. \\ \hline

	\end{tabular}
}
\end{table}

\bigskip
The real datasets employed are: the \muacro{ISOMAP} face database~\cite{Tenenbaum2000}, the \muacro{MNIST} database \cite{Lecun1998}, 
the \muacro{Santa Fe}~\cite{Pineda1994} dataset, the \muacro{Isolet} dataset~\cite{Isolet}, and the \muacro{DSVC1} time series~\cite{Camastra2009}.

The \muacro{ISOMAP} face database consists in $698$ gray-level images of size $64\times64$ depicting the face of a sculpture. 
This dataset has three degrees of freedom: two for the pose and one for the lighting direction. 

The \muacro{MNIST} database consists in $70000$ gray-level images of size $28\times28$ of hand-written digits; 
in our tests we used the $6742$ training points 
representing the digit $1$.
The \muacro{id} of this database is not actually known; we therefore rely on the estimations proposed in~\cite{Hein2005,Costa2005} for the different digits, and in particular on the range $\{8..11\}$ for the digit $1$.

The version $D2$ of the \muacro{Santa Fe} dataset is a synthetic time series of $50000$ one-dimensional points; it was generated by a simulation of particle motion, and it has nine degrees of freedom.
In order to estimate the attractor dimension of this time series, we used the method of delays described in~\cite{Ott1993}, which generates $D$-dimensional vectors by collecting $D$ values from the original dataset; by choosing $D=50$ we obtained a dataset containing $1000$ points in $\Re^{50}$.

The \muacro{Isolet} dataset has been generated as follows: $150$ subjects spoke the name of each letter of the alphabet twice, thus producing $52$ training examples from each speaker. The latter are grouped into sets of $30$ speakers each, and are referred to as $isolet1$, $isolet2$, $isolet3$, $isolet4$, and $isolet5$, for a total of $7797$ samples. The \muacro{id} of this dataset is not actually known, 
but a study reported in~\cite{Kivimaki2010} has proposed that the correct estimation could be in the range $\{16..22\}$.

The \muacro{DSVC1} is a real data time series composed of $5000$ samples and measured from a hardware realization of the 
Chua's circuit~\cite{Chua1985}. We used the method of delays choosing $D=20$, and we obtained a dataset containing $250$ points in $\Re^{20}$; the \muacro{id} of this dataset is $\sim 2.26$ as reported in~\cite{Camastra2009}.

\subsection{Experimental Setting}\seclabel{expSettings}

To objectively assess our method, we compared it with well-known \muacro{id} estimators such as: \muacro{SPPCA},  
\muacro{kNNG}, \muacro{CD}, \muacro{MLE}, \muacro{Hein}, \muacro{BPCA}, $\muacro{MiND}_\muacro{KL}$, and \muacro{IDEA}.
For \muacro{kNNG}, \muacro{MLE}, \muacro{Hein}, \muacro{BPCA}, $\muacro{MiND}_\muacro{KL}$, and \muacro{IDEA} we used the authors' implementation%
\footnote{\scriptsize{http://www.eecs.umich.edu/$\sim$hero/IntrinsicDim/, 
					\\http://www.stat.lsa.umich.edu/$\sim$elevina/mledim.m,
					\\http://www.ml.uni-saarland.de/code.shtml,
					\\http://research.microsoft.com/en-us/um/cambridge/projects/infernet/blogs/bayesianpca.aspx
					\\http://security.dico.unimi.it/$\sim$fox721/}}, while for the other algorithms we employed the version provided by the dimensionality reduction toolbox%
\footnote{\scriptsize{http://cseweb.ucsd.edu/$\sim$lvdmaaten/dr/download.php}}. 

To generate the synthetic datasets we adopted the modified generator described in~\secref{datasets} creating $20$ instances of each dataset 
reported in~\tabref{synthDatasets}, each of which is composed by $2500$ randomly sampled points.
 
To obtain an unbiased estimation, for each technique we averaged the results achieved on the $20$ instances.
To execute multiple tests also on $\VSP{M}_\muacro{MNIST1}$ and $\VSP{M}_\muacro{Isolet}$ we extracted $5$ random subsets containing $2500$ points each, and we 
averaged the achieved results.

\begin{table}[!t]
	\centering
	\caption{Parameter settings for the different estimators: $k$ 
		  represents the number of neighbors, $\gamma$ is the edge weighting factor for \muacro{kNNG}, $M$ is the number of Least Square (\muacro{LS}) 
		  runs, $N$ is the number of resampling trials per \muacro{LS} iteration, 
		  $\alpha$ and $\pi$ represent the parameters (shape and rate) of the Gamma prior distributions describing 
		  the hyper-parameters and the observation noise model of $\muacro{BPCA}$, 
		  $\mu$ contains the mean and the precision of the Gaussian prior distribution 
		  describing the bias inserted in the inference of $\muacro{BPCA}$.
		}
	\tablabel{parameters}
{
	\begin{tabular}{|c|c|c|}
		\hline
		\textbf{Dataset} &\textbf{Method} & \textbf{Parameters} \\
		\hline
		\hline
		\multirow{10}{*}{\textbf{Synthetic}} 
		&\muacro{SPPCA} & $None$\\
		&\muacro{CD} & $None$\\ 
		&\muacro{MLE} & $k_1 = 6$ $k_2 = 20$\\ 
		&$\muacro{kNNG}_{1}$ & $k_1 = 6, k_2 = 20, \gamma = 1, M=1, N=10$\\ 
		&$\muacro{kNNG}_{2}$ & $k_1 = 6, k_2 = 20, \gamma = 1, M=10, N=1$\\ 
		&$\muacro{BPCA}$ & $iters = 500$, $\alpha = (2.0, 2.0)$ $\pi = (2.0, 2.0)$ $\mu = (0.0, 0.01)$\\
		&$\muacro{MiND}_\muacro{KL}$ & $k = 10$ \\ 
		&$\muacro{IDEA}$ & $k = 10$ \\ 
		&$\muacro{DANCo}$ & $k = 10$ \\ \hline \hline
		\multirow{10}{*}{\textbf{Real}}
		&\muacro{SPPCA} & $None$\\
		&\muacro{CD} & $None$\\ 
		&\muacro{MLE}  & $k_1 = 3$ $k_2 = 8$\\ 
		&$\muacro{kNNG}_{1}$ & $k_1 = 3, k_2 = 8, \gamma = 1, M=1, N=10$\\ 
		&$\muacro{kNNG}_{2}$ & $k_1 = 3, k_2 = 8, \gamma = 1, M=10, N=1$\\ 
		&$\muacro{BPCA}$ & $iters = 2000$, $\alpha = (2.0, 2.0)$ $\pi = (2.0, 2.0)$ $\mu = (0.0, 0.01)$\\
		&$\muacro{MiND}_\muacro{KL}$ & $k = 5$ \\ 
		&$\muacro{IDEA}$ & $k = 5$ \\
		&$\muacro{DANCo}$ & $k = 5$ \\ \hline
	\end{tabular}
}
\end{table}
In \tabref{parameters} the configuration parameters employed in our tests are summarized.
To relax the dependency of the \muacro{kNNG} algorithm from the selection of the value of its parameter $k$, we performed multiple runs 
with $k_1 \leq k \leq k_2$ and we averaged the achieved results (see~\tabref{parameters}).

\subsection{Experimental Results}\seclabel{results}

This section reports the results achieved on both synthetic and real datasets. 
In particular, \tabref{res1} summarizes the results obtained on the synthetic datasets. 
\begin{table}[!t]

 	\centering
 	\caption{Results achieved on the synthetic datasets. The best approximations are highlighted in boldface.}
 	\tablabel{res1}
{
 	\begin{tabular}{|c|c||c|c|c|c|c|c|c|c|c|c|}

 	 \hline
 	  Dataset & $d$ & \muacro{SPPCA} &\muacro{BPCA} & $\muacro{kNNG}_{1}$ & $\muacro{kNNG}_{2}$ & \muacro{CD} & \muacro{MLE} & \muacro{Hein} & $\muacro{MiND}_\muacro{KL}$ & \muacro{IDEA} & \muacro{DANCo}\\
 	  \hline
 	  \hline
 $\VSP{M}_{12}$ & $\textit{1}$  & $3.00$ & $5.70$ & $0.97$ & $1.07$ & $1.14$ & $\mathbf{1.00}$ & $\mathbf{1.00}$ & $\mathbf{1.00}$ & $1.02$ & $\mathbf{1.00}$ \\  
 $\VSP{M}_5$ 	 & $\textit{2}$ & $3.00$ & $\mathbf{2.00}$ & $1.96$ & $2.06$ & $1.98$ & $1.97$ & $\mathbf{2.00}$ & $\mathbf{2.00}$ & $\mathbf{2.00}$ & $\mathbf{2.00}$  \\ 
 $\VSP{M}_7$ 	 & $\textit{2}$ & $3.00$ & $\mathbf{2.00}$ & $1.97$ & $2.09$ & $1.93$ & $1.96$ & $\mathbf{2.00}$ & $\mathbf{2.00}$ & $2.07$ & $\mathbf{2.00}$  \\ 
 $\VSP{M}_{10}$ & $\textit{2}$  & $3.00$ & $1.55$ & $1.95$ & $2.03$ & $2.19$ & $2.21$ & $\mathbf{2.00}$ & $\mathbf{2.00}$ & $1.98$ & $\mathbf{2.00}$  \\ 
 $\VSP{M}_2$ 	 & $\textit{3}$ & $\mathbf{3.00}$ & $\mathbf{3.00}$ & $2.95$ & $3.03$ & $2.88$ & $2.88$ & $\mathbf{3.00}$ & $\mathbf{3.00}$ & $3.03$ & $\mathbf{3.00}$  \\ 
 $\VSP{M}_3$ 	 & $\textit{4}$ & $\mathbf{4.00}$ & $\mathbf{4.00}$ & $3.75$ & $3.82$ & $3.23$ & $3.83$ & $\mathbf{4.00}$ & $\mathbf{4.00}$ & $4.01$ & $\mathbf{4.00}$  \\ 
 $\VSP{M}_4$ 	 & $\textit{4}$ & $8.00$ & $4.25$ & $4.05$ & $4.76$ & $3.88$ & $3.95$ & $\mathbf{4.00}$ & $4.15$ & $3.93$ & $\mathbf{4.00}$  \\ 
 $\VSP{M}_6$    & $\textit{6}$  & $12.00$& $12.00$& $6.46$ & $11.24$& $5.91$ & $6.39$ & $\mathbf{5.95}$ & $6.50$ & $6.33$ & $6.90$  \\ 
 $\VSP{M}_1$ 	 & $\textit{10}$& $11.00$& $5.45$ & $9.16$ & $9.89$ & $9.12$ & $9.10$ & $9.45$ & $10.30$& $10.41$& $\mathbf{10.00}$  \\ 
 $\VSP{M}_{9a}$ & $\textit{10}$ & $\mathbf{10.00}$& $5.20$ & $8.62$ & $10.21$& $8.09$ & $8.26$ & $8.90$ & $9.85$ & $9.93$ &  
$9.50$  \\ 
 $\VSP{M}_{9b}$ & $\textit{17}$ & $\mathbf{17.00}$& $9.46$ & $13.69$& $15.38$& $12.30$& $12.87$& $13.85$& $16.25$& $16.07$& $16.47$  \\ 
 $\VSP{M}_{13}$ & $\textit{18}$ & $36.00$& $36.00$& $17.58$& $5.01$ & $11.60$& $15.95$& $14.00$& $18.60$& $17.30$& $\mathbf{18.20}$  \\ 
 $\VSP{M}_{8}$ & $\textit{20}$ & $\mathbf{20.00}$& $13.55$& $15.25$& $10.59$& $13.75$& $14.64$& $15.50$& $19.15$& $18.51$& $19.54$  \\ 
 $\VSP{M}_{11}$ & $\textit{20}$ & $\mathbf{20.00}$& $13.70$& $16.40$& $24.89$& $11.26$& $15.82$& $15.00$& $19.35$& $21.20$& $19.90$  \\ 
 $\VSP{M}_{9c}$ & $\textit{24}$ & $\mathbf{24.00}$& $13.3$ & $17.67$& $21.42$& $15.58$& $16.96$& $17.95$& $22.55$& $23.93$& $23.85$  \\ 
 $\VSP{M}_{14}$ & $\textit{24}$ & $48.00$& $48.00$& $19.66$& $22.80$& $14.03$& $19.83$& $17.00$& $25.30$& $22.90$& $\mathbf{25.00}$  \\ 
 $\VSP{M}_{9d}$	& $\textit{70}$ & $71.00$& $71.00$ & $39.67$&$40.31$& $31.4$ & $36.49$& $38.69$& $65.30$& $46.7$ & $\mathbf{70.42}$ \\ 
		\hline
 	  \muacro{MPE} & & $44.79$ & $61.55$ & $11.72$ & $20.14$ & $20.79$ & $13.78$ & $12.04$ & $2.94$ & $4.75$ &  $\mathbf{1.90}$ \\ \hline
 	\end{tabular}
}
\end{table}
It is possible to note that the best performing algorithm is \muacro{DANCo}. Indeed, this estimator can correctly deal with linear and nonlinear manifolds embedded in low and high dimensional spaces.
In particular, it is the only method that achieves a good estimation for the three datasets $\VSP{M}_{9d}$, $\VSP{M}_{13}$, and $\VSP{M}_{14}$.

Instead, geometrical approaches, such as $\muacro{kNNG}$, \muacro{CD}, \muacro{MLE}, and \muacro{Hein}, obtain good estimates only for low \muacro{id} manifolds, failing to deal with high \muacro{id} data. Moreover, the projection techniques, such as \muacro{BPCA} and \muacro{SPPCA}, are able to correctly deal only with linear embedded manifolds. These considerations confirm that the geometric methods are affected by an underestimation bias as noticed in~\cite{ECML2011,ICIAP2011} and that all the projection methods cannot provide reliable \muacro{id} estimates~\cite{Levina2005}. 

Furthermore,  \muacro{DANCo} outperforms also \muacro{IDEA} and $\muacro{MiND}_\muacro{KL}$ that have been developed to deal with datasets having a sufficiently high \muacro{id} (that is \muacro{id} $\geqslant 10$) and being drawn from manifolds nonlinearly embedded in higher dimensional spaces. 

In the last row of \tabref{res1} the Mean Percentage Error (\muacro{MPE}) indicator, proposed in~\cite{ECML2011} in order to evaluate the overall performance of a given estimator, is reported. 
For each algorithm this value is computed as the mean of the percentage errors obtained on each dataset, i.e. 
$		\muacro{MPE} = \frac{100}{\#\VSP{M}}\sum_{\VSP{M}} \frac{|\hat{d}_{\VSP{M}}-d_{\VSP{M}}|}{d_{\VSP{M}}}$, 
where $d_{\VSP{M}}$ is the real \muacro{id}, $\hat{d}_{\VSP{M}}$ is the estimated one, and $\#\VSP{M}$ is the number of tested manifolds. Considering this indicator, \muacro{DANCo} ranks as the best performing estimator. 

\begin{table}[!t]
 	\centering
 	\caption{Results achieved on the real datasets by the employed approaches. 
		The best approximations are highlighted in boldface. }
 	\tablabel{res2}
{
\begin{tabular}{|c|c||c|c|c|c|c|c|c|c|c|c|}
 	 \hline
 	  Dataset & $d$ & \muacro{SPPCA} &\muacro{BPCA} & $\muacro{kNNG}_{1}$ & $\muacro{kNNG}_{2}$ & \muacro{CD} & \muacro{MLE} & \muacro{Hein} & $\muacro{MiND}_\muacro{KL}$ & \muacro{IDEA} & \muacro{DANCo}\\
 	  \hline
 	 \hline
 	  $\VSP{M}_\muacro{DSVC1}$ 	& $\textit{2.26}$ & $4.00$ & $6.00$ & $1.77$ & $1.86$ & $1.92$ & $2.03$ & $3.00$ & $2.50$ & $2.14$ & $\mathbf{2.26}$\\ 
	  $\VSP{M}_\muacro{Faces}$ 	& $\textit{3}$& $5.00$ & $4.00$ & $3.60$ & $4.32$ & $3.37$ & $4.05$ & $\mathbf{3.00}$ & $3.90$ & $3.73$ & $4.00$\\ 
	  $\VSP{M}_\muacro{Santa Fe}$ 	& $\textit{9}$& $19.00$ & $18.00$ & $7.28$ & $7.43$ & $4.39$ & $7.16$ & $6.00$ & $7.60$ & $7.26$ & $\mathbf{8.19}$\\ 
          $\VSP{M}_\muacro{MNIST1}$ 	& $\textit{8-11}$& $9.00$ & $11.00$ & $10.37$ & $\mathbf{9.58}$ & $6.96$ & $10.29$ & $8.00$ & $11.00$ & $11.06$ & $9.98$\\ 
	  $\VSP{M}_\muacro{Isolet}$ 	& $\textit{16-22}$& $45.00$ & $\mathbf{19.00}$ & $6.50$ & $8.32$ & $3.65$ & $15.78$ & $3.00$ & $20.00$ & $18.77$ & $\mathbf{19.00}$ \\ \hline 
	  \muacro{MPE} 		    	& & $79.37$ & $62.92$ & $27.14$ & $27.24$ & $37.22$ & $18.17$ & $33.21$ & $15.44$ & $13.32$ & $\mathbf{9.47}$\\ \hline
	  
	  \end{tabular}
}
\end{table}
In~\tabref{res2} the results obtained on real datasets are summarized. 
Being the real data generally affected by the presence of noise, the quality of the estimates computed by the projection methods is strongly reduced, as confirmed by the poor results obtained by \muacro{BPCA} and \muacro{SPPCA}. 
The geometric approaches we tested are less affected by noise, but they are not able to correctly deal with the high dimensionality of the $\VSP{M}_\muacro{Isolet}$ dataset.

As can be seen, \muacro{DANCo} is the best performing estimator, strongly overcoming also the results obtained by those techniques, such as \muacro{IDEA} and $\muacro{MiND}_\muacro{KL}$, that exploit a correction approach.
These results, together with the best average estimation precision achieved by our technique in terms of \muacro{MPE}\footnote{Where the true value of the \muacro{id} is not known, we considered the mean value of the range as $d_{\VSP{M}}$.}, confirm that \muacro{DANCo} is a promising and valuable tool for \muacro{id} estimation. 

Finally, to test the robustness of our algorithms w.r.t. the choice of the parameter $k$, 
we employed \muacro{DANCo} to reproduce the experiments proposed for the \muacro{MLE} algorithm in Figure~1~(a) of~\cite{Levina2005} and in Figure~2 of~\cite{ECML2011}, and we averaged the curves obtained in $10$ runs. 
In these tests the adopted datasets are composed by points drawn from the standard Gaussian \muacro{pdf} in $\Re^5$. 
We repeated the test for datasets with cardinalities $N\in\{200,500,1000,2000\}$ varying the parameter $k$ in the range $\{5..100\}$. 
For all the combinations of the dataset cardinalities and the $k$ parameter values, \muacro{DANCo} obtained $\muacro{id}$ estimates always equal to $5$, confirming its strong robustness.

\section{Conclusions and Future Works}\seclabel{Conclusions}

In this paper we proposed a novel consistent estimator, called \muacro{DANCo}, that combines the effects of concentration of angles and norms to estimate the \muacro{id} of a given dataset.
The proposed method compares the joint \muacro{pdf} estimated on the dataset, related to angles and norms respectively, with those computed on synthetic datasets of known \muacro{id}; to this aim, a closed-form expression for the Kullback-Leibler divergence of their distributions is employed. 

We tested our algorithm on both synthetic and real datasets comparing its results with those obtained by employing well-known \muacro{id} estimators. 
The overall results show that \muacro{DANCo} is a really promising and valuable technique for \muacro{id} estimation. Indeed, it provides the most accurate results, computing either the best \muacro{id} estimates or values that are strongly comparable to the best ones. Moreover, this algorithm has shown to be really robust in terms of its capability to: i) deal with both high and low \muacro{id}, ii) manage both linearly and nonlinearly embedded manifolds, and iii) outperform all the other estimators on noisy real datasets.

\par\bigskip
Future works will be devoted to identify a bound for the finite sample error,
to further formally evaluate the effectiveness of the proposed approach.


{

\begin{thebibliography}{10}

\bibitem{Bishop1998}
C.M. Bishop.
\newblock Bayesian {PCA}.
\newblock {\em Proc. of NIPS}, 11:382--388, 1998.

\bibitem{Breitenberger1963}
E.~Breitenberger.
\newblock Analogues of the normal distribution on the circle and the sphere.
\newblock {\em Biometrika}, 50, 1963.

\bibitem{Camastra2003}
F.~Camastra.
\newblock Data dimensionality estimation methods: A survey.
\newblock {\em Pattern Recognition}, 36(12):2945--2954, 2003.

\bibitem{Camastra2009}
F.~Camastra and M.~Filippone.
\newblock A comparative evaluation of nonlinear dynamics methods for time
  series prediction.
\newblock {\em Neural Computing and Applications}, 18(8):1021--1029, November
  2009.

\bibitem{Camastra2002}
F.~Camastra and A.~Vinciarelli.
\newblock Estimating the intrinsic dimension of data with a fractal-based
  method.
\newblock {\em IEEE Trans. on PAMI}, 24:1404--1407, 2002.

\bibitem{Chua1985}
L.~Chua, M.~Komuro, and T.~Matsumoto.
\newblock The double scroll.
\newblock {\em IEEE Trans. on Circuits and Systems}, 32:797--818, 1985.

\bibitem{Costa2004b}
J.A. Costa and A.O. Hero.
\newblock Geodesic entropic graphs for dimension and entropy estimation in
  manifold learning.
\newblock {\em IEEE Trans. on Signal Processing}, 52(8):2210--2221, 2004.

\bibitem{Costa2004}
J.A. Costa and A.O. Hero.
\newblock Learning intrinsic dimension and entropy of high-dimensional shape
  spaces.
\newblock In {\em Proc. of EUSIPCO}, 2004.

\bibitem{Costa2005}
J.A. Costa and A.O. Hero.
\newblock Learning intrinsic dimension and entropy of shapes.
\newblock In {\em Stat. and anal. of shapes}. Birkhauser, 2005.

\bibitem{Eckmann1992}
J.P. Eckmann and D.~Ruelle.
\newblock Fundamental limitations for estimating dimensions and {L}yapunov
  exponents in dynamical systems.
\newblock {\em Physica D: Nonlinear Phenomena}, 56(2-3):185--187, 1992.

\bibitem{Farahmand2007}
A.M. Farahmand, C.~Szepesvari, and J.Y. Audibert.
\newblock Manifold-adaptive dimension estimation.
\newblock {\em Proc. of ICML}, 2007.

\bibitem{Fisher1981}
N.~I. Fisher.
\newblock {\em {Statistical Analysis of Circular Data}}.
\newblock Cambridge University Press, January 1996.

\bibitem{Isolet}
A.~Frank and A.~Asuncion.
\newblock {UCI} machine learning repository, 2010.

\bibitem{Friedman2009}
J.H. Friedman, T.~Hastie, and R.~Tibshirani.
\newblock {\em The {E}lements of {S}tatistical {L}earning - {D}ata {M}ining,
  {I}nference and {P}rediction}.
\newblock Springer, Berlin, 2009.

\bibitem{Fukunaga1971}
K.~Fukunaga.
\newblock An algorithm for finding intrinsic dimensionality of data.
\newblock {\em IEEE Trans. on Computers}, 20:176--183, 1971.

\bibitem{Fukunaga1982}
K.~Fukunaga.
\newblock {\em {I}ntrinsic {D}imensionality {E}xtraction}.
\newblock Classification, Pattern Recognition and Reduction of Dimensionality.
  P.R. Krishnaiah and L.N. Kanal, Amsterdam: North Holland, 1982.

\bibitem{Grassberger1983}
P.~Grassberger and I.~Procaccia.
\newblock Measuring the strangeness of strange attractors.
\newblock {\em Physica D: Nonlinear Phenomena}, 9:189--208, 1983.

\bibitem{Guan2009}
Y.~Guan and J.~G. Dy.
\newblock Sparse probabilistic principal component analysis.
\newblock {\em J. of Machine Learning Research - Proc. Track}, 5:185--192,
  2009.

\bibitem{Hein2005}
M.~Hein.
\newblock Intrinsic dimensionality estimation of submanifolds in euclidean
  space.
\newblock In {\em Proc. of ICML}, pages 289--296, 2005.

\bibitem{Hill1976}
G.~W. Hill.
\newblock New approximations to the von {M}ises distribution.
\newblock {\em Biometrika}, 63(3):673--676, 1976.

\bibitem{Bellman1961}
I.T. Jollife.
\newblock {\em Adaptive {C}ontrol {P}rocesses: {A} {G}uided {T}our}.
\newblock Princeton University Press, 1961.

\bibitem{Jollife1986}
I.T. Jollife.
\newblock {\em {P}rincipal {C}omponent {A}nalysis}.
\newblock Springer Series in Statistics. Springer-Verlag, New York, NY, 1986.

\bibitem{Kirby2001}
M.~Kirby.
\newblock {\em Geometric {D}ata {A}nalysis: an {E}mpirical {A}pproach to
  {D}imensionality {R}eduction and the {S}tudy of {P}atterns}.
\newblock John Wiley and Sons, 1998.

\bibitem{Kivimaki2010}
I.~Kivim\"{a}ki, K.~Lagus, I.~Nieminen, J.~V\"{a}yrynen, and T.~Honkela.
\newblock Using correlation dimension for analysing text data.
\newblock In {\em Proc. of the ICANN}, pages 368--373, Berlin, Heidelberg,
  2010. Springer-Verlag.

\bibitem{Lecun1998}
Y.~LeCun, L.~Bottou, Y.~Bengio, and P.~Haffner.
\newblock Gradient-based learning applied to document recognition.
\newblock {\em Proc. of IEEE}, 86:2278--2324, 1998.

\bibitem{Levina2005}
E.~Levina and P.J. Bickel.
\newblock Maximum likelihood estimation of intrinsic dimension.
\newblock {\em Proc. of NIPS 17}, 1:777--784, 2005.

\bibitem{Li2010}
J.~Li and D.~Tao.
\newblock Simple exponential family {PCA}.
\newblock {\em Proc. of AISTATS}, pages 453--460, 2010.

\bibitem{ECML2011}
G.~Lombardi, A.~Rozza, C.~Ceruti, E.~Casiraghi, and P.~Campadelli.
\newblock {M}inimum neighbor distance estimators of intrinsic dimension.
\newblock {\em Proc. of ECML-PKDD}, 6912:374--389, 2011.

\bibitem{Lord1954}
R.~D. Lord.
\newblock The use of the {H}ankel transform in statistics {I}. general theory
  and examples.
\newblock {\em Biometrika}, 41(1/2):44--55, 1954.

\bibitem{Mardia1972}
K.~V. Mardia.
\newblock {\em {Statistics of Directional Data}}.
\newblock Academic Press, 1972.

\bibitem{Ott1993}
E.~Ott.
\newblock {\em Chaos in Dynamical Systems}.
\newblock Cambridge University Press, Cambridge, 1993.

\bibitem{Pineda1994}
F.J. Pineda and J.C. Sommerer.
\newblock Estimating generalized dimensions and choosing time delays: A fast
  algorithm.
\newblock {\em Time Series Prediction. Forecasting the Future and Understanding
  the Past}, pages 367--385, 1994.

\bibitem{journalMlj}
A.~Rozza, G.~Lombardi, C.Ceruti, E.~Casiraghi, and P.~Campadelli.
\newblock Novel high intrinsic dimensionality estimators.
\newblock {\em Machine Learning J.}, May 2012.

\bibitem{ICIAP2011}
A.~Rozza, G.~Lombardi, M.~Rosa, E.~Casiraghi, and P.~Campadelli.
\newblock {IDEA}: Intrinsic dimension estimation algorithm.
\newblock {\em Proc. ICIAP}, 6978:433--442, 2011.

\bibitem{Sodergren2011}
A.~Sodergren.
\newblock On the distribution of angles between the {N} shortest vectors in a
  random lattice.
\newblock {\em J. London Math. Soc.}, 84(3):749--764, 2011.

\bibitem{Tenenbaum2000}
J.~Tenenbaum, V.~Silva, and J.~Langford.
\newblock A global geometric framework for nonlinear dimensionality reduction.
\newblock {\em Science}, 290:2319--2323, 2000.

\bibitem{Upton1986}
G.~J.~G. Upton.
\newblock Approximate confidence intervals for the mean direction of a von
  {M}ises distribution.
\newblock {\em Biometrika}, 73(2):525--527, 1986.

\bibitem{Vapnik1998}
V.~Vapnik.
\newblock {\em Statistical {L}earning {T}heory}.
\newblock John Wiley and Sons, 1998.

\bibitem{Verveer1995}
P.J. Verveer and R.P.W. Duin.
\newblock An evaluation of intrinsic dimensionality estimators.
\newblock {\em IEEE Trans. on PAMI}, 17:81--86, 1995.

\bibitem{Oraintara2008}
A.P.N. Vo, S.~Oraintara, and T.T. Nguyen.
\newblock Statistical image modeling using von {M}ises distribution in the
  complex directional wavelet domain.
\newblock In {\em Proc. of ISCAS 2008}, pages 2885--2888, 2008.

\bibitem{Wang2006}
Q.~Wang, S.R. Kulkarni, and S.~Verdú.
\newblock A nearest-neighbor approach to estimating divergence between
  continuous random vector.
\newblock In {\em Proc. ISIT}, pages 242--246, 2006.

\end{thebibliography}
}


\end{document}